\documentclass{article}
\usepackage{subfig}

\usepackage{PRIMEarxiv}

\usepackage[utf8]{inputenc} 
\usepackage[T1]{fontenc}    
\usepackage{hyperref}       
\usepackage{url}            
\usepackage{booktabs}       
\usepackage{amsfonts}       
\usepackage{nicefrac}       
\usepackage{microtype}      
\usepackage{lipsum}
\usepackage{fancyhdr}       
\usepackage{graphicx}       
\graphicspath{{media/}}     

\title{Beyond Next Word Prediction: Developing Comprehensive Evaluation
Frameworks for measuring LLM performance on real world applications
}

\author{
  Vishakha Agrawal \\
  AMD \\
  \texttt{vishakha.research.id@gmail.com} \\
   \And
  Archie Chaudhury \\
  LayerLens\\
  \texttt{ac@layerlens.ai} \\
  \And
  Shreya Agrawal\\
  UCLA\\
  agrawalshreya@g.ucla.edu\\
}

\begin{document}
\maketitle

\begin{abstract}
While Large Language Models (LLMs) are fundamentally next-token prediction systems, their practical applications extend far beyond this basic function. From natural language processing and text generation to conversational assistants and software use, LLMs have numerous use-cases, and have already acquired a significant degree of enterprise adoption. To evaluate such models, static evaluation datasets, consisting of a set of prompts and their corresponding ground truths, are often used to benchmark the efficacy of the model for a particular task. In this paper, we provide the basis for a more comprehensive evaluation framework, based upon a traditional game and tool-based architecture that enables a more overarching measurement of a model's capabilities. For simplicity, we provide a generalized foundation that can be extended, without significant alteration, to numerous scenarios, from specific use cases such as supply chain management or financial reasoning, to abstract measurements such as ethics or safety.  
\end{abstract}


\section{Introduction}
The emergence of Large Language Models (LLMs) has fundamentally transformed our understanding of artificial intelligence capabilities. Although these models are, at their core, sophisticated next-token prediction systems \cite{zellers2019hellaswag}, they have been adopted for use-cases that far transcend this fundamental architecture, forming the basis for an ubiquitous agent-based economy in which LLMs are entrusted with making key decisions that can have significant financial, ethical, or moral consequences\cite{zhao2023survey}. In fact, LLMs are already being used to automate the development of software or create authorized legal documents. This apparent disconnect between their basic function and advanced capabilities creates significant challenges in evaluating their suitability and practical performance. As organizations increasingly look to deploy LLMs in various applications, the need for sophisticated evaluation frameworks becomes critical.

\section{Current State of LLM Evaluations}
Evaluations, in the context of LLMs, refer to the testing of models on specific use cases or tasks, often conducted after training or fine-tuning. Evaluations often vary in their overall complexity: an evaluation can consist of a constrained set of static question answer pairs with easily verifiable answers, or a multi-turn environment, which often requires outfitting the LLM with a specific set of tools and definitions, and giving it access to external scaffolding which can execute API calls created by the LLM against the environment. Examples of the former include Massive Multimodal Language Understanding (MMLU)\cite{chen2024evolution}, General Purpose Question Answering (GPQA)\cite{rein2024gpqa}, and BigBenchHard\cite{suzgun2022challenging}. In all of these cases, the LLM is prompted individually with a question and is expected to output an answer that adheres to some predefined format for verification. Different types of scoring functions often exist for static datasets: questions can either be verified through performing a computational string match, or by leveraging an external judge, which can either be another language model or human annotator.  For environment-based benchmarks, examples include SWE Bench\cite{jimenez2023swe}, Berkeley Function Calling (BFCL) \cite{patil2023gorilla}, and SWE-Lancer \cite{miserendino2025swe}; these are all evaluations where the answer is not known apriori, and an LLM is instructed to generate executable functions, in the form of code or function calls, to solve a particular task. Tasks are scored by either automatically verifying the output on the environment, or again, having an external judge evaluate the output according to some predefined criteria.

\section{Model Specialization and Task Suitability}
Recent benchmarking data has revealed a fascinating pattern in LLM capabilities: different models excel at distinctly different types of tasks, despite sharing similar fundamental architectures. This specialization appears to emerge from various factors including training approaches, model architecture decisions, and optimization choices.\newline
Consider the case of DeepSeek's R1 model, which has demonstrated exceptional capabilities in mathematical reasoning and analytical tasks. According to LayerLens Atlas benchmarking data, R1 outperforms many competitors in areas requiring deep analytical thinking, particularly in mathematics and stock market analysis. However, it shows relatively weaker performance in practical business tasks such as spreadsheet management and bookkeeping. This specialization appears to be a direct result of R1's architecture, which favors thorough analytical processing at the cost of speed.\newline

\begin{figure}
\subfloat{\includegraphics[width = 7in]{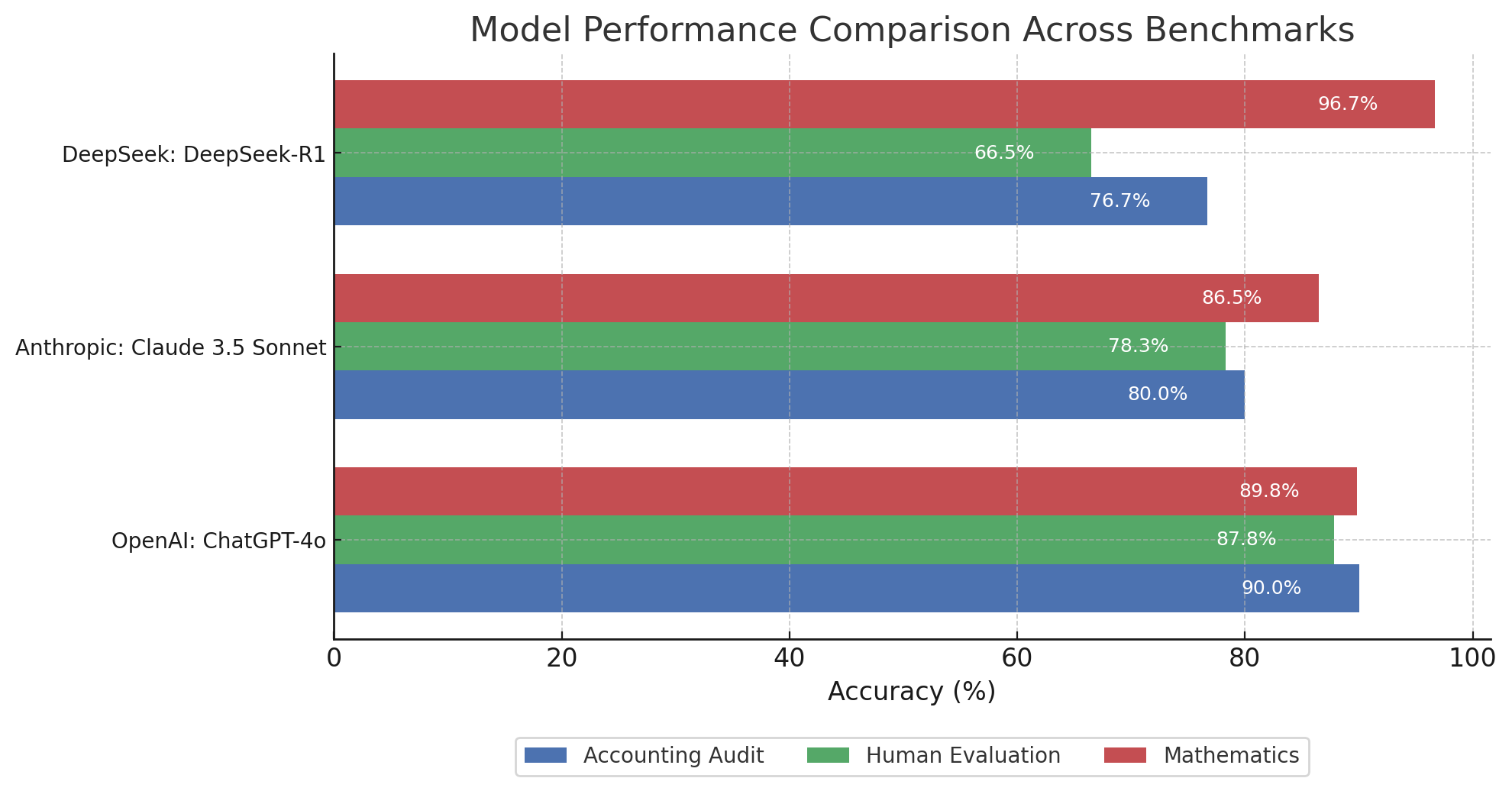}}
\caption{Performance of models on financial reasoning, Coding and Maths}
\label{rcm}
\end{figure}

\begin{figure}
\subfloat{\includegraphics[width = 7in]{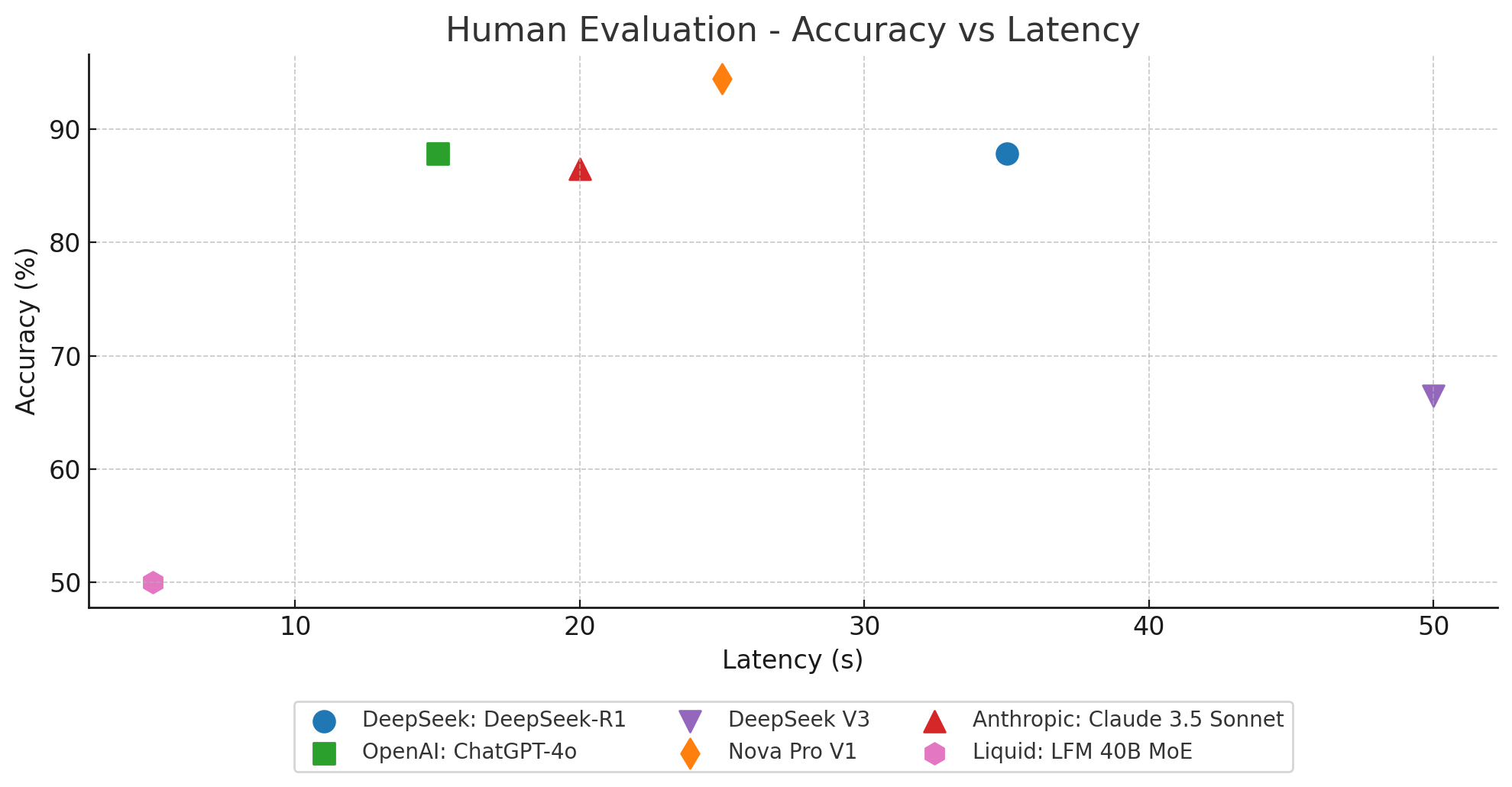}} 
\caption{Accuracy vs Latency on HumanEval, a dataset measuring ability to write code }
\label{latency}
\end{figure}

\begin{figure}
\subfloat{\includegraphics[width = 7in]{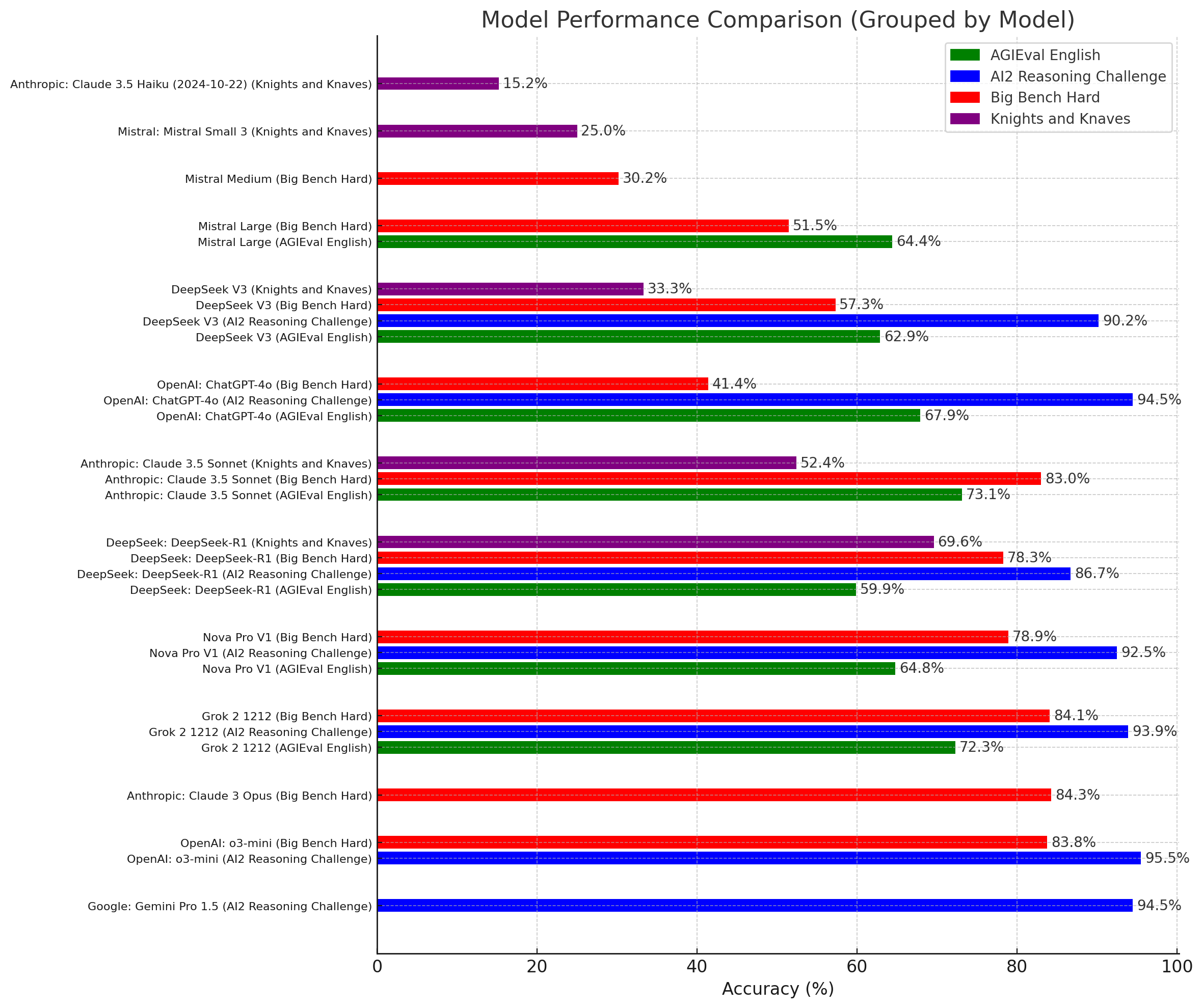}} 
\caption{Performance of different models across different datasets}
\label{more}
\end{figure}

The data cited above may differ from official measurements and was calculated independently through the LayerLens Atlas application, which allows for the independent benchmarking of foundational models. We used the Mathematics\cite{hendrycks2021measuring} dataset to measure mathematical reasoning, Humanity's Last Exam for natural sciences, \cite{phan2025humanitysexam} , Knight and Knaves\cite{xie2024memorization} for logical reasoning, and HumanEval\cite{chen2024evolution} for computer science and programming. BigBench Hard\cite{suzgun2022challenging} and AGIEVal\cite{zhong2023agieval} are used to measure general knowledge and English comprehension skills, respectively:   \ref{rcm}, \ref{latency}, \ref{more},.

In contrast, Anthropic's Claude\cite{claude}, \cite{claude2} models have established themselves as leaders in code generation and structured tasks. Our results show Claude consistently achieving superior results in generating runnable Python code, though interestingly, this advantage appears limited to functional correctness rather than code efficiency or style. This specialization in structured, rule-based tasks does not indicate performance in contemporary, real world-scenarios, and instead show a preference for systematic tasks.

OpenAI's GPT models present yet another specialization pattern, showing particular strength in practical business applications. These models excel in tasks requiring structured, rule-based reasoning such as spreadsheet management and enterprise automation. This capability likely stems from extensive training on business-related datasets and optimization for practical applications.\newline
The emergence of these specializations has important implications for both model development and deployment decisions. Organizations must carefully consider these patterns when selecting models for specific applications. A model that excels at mathematical analysis might not be the best choice for customer service applications, while a model optimized for code generation might not be ideal for creative writing tasks.\newline
These specializations also reveal interesting insights about the nature of LLM capabilities. Despite being fundamentally next-token prediction systems, different models develop distinct strengths and weaknesses that mirror human specialization patterns. This suggests that the way models are trained and optimized can lead to the emergence of specialized capabilities, much like human expertise develops through focused practice and study.

Furthermore, these specializations also signal a difference between models optimized for objective, yet limited tasks, and models optimized for real-world use-cases. Indeed, there are significant limitations in the capabilites of current evaluations to accurately measure the ability of models to perform well on practical tasks. 

\section{Limitations of current evaluation datasets}
The limitations of current evaluation methods extend beyond their narrow focus. Many benchmarks fail to account for crucial factors such as response time, resource utilization, and user experience. The case of R1 demonstrates this clearly, while the model excels in certain analytical tasks, its slower processing speed impacts its viability for time-sensitive applications. This reveals a critical gap in current evaluation frameworks that must be addressed.

The proliferation of new models outpaces the rate of creation for new benchmarks. New model releases often report their performance on the same constrained set of benchmarks, often reaching state-of-the-art (SOTA) performance on them within a couple of releases. The current suite of benchmarks, often concentrated on reasoning or knowledge-based tasks, also deviate significantly from the manner in which the majority of enterprises and end-users interact with LLMs. For example, the AIME dataset\cite{oneaime}, which has often been cited as an indicator of performance in the release of new models such as Grok 3, is a collection of questions from the AIME mathematical competition: while this may be an interesting barometer to measure theoretical progress, it is ultimately irrelevant to a business or user who may want to use the model to automate underlying processes. \newline

Furthermore, current benchmarks (both static and environment variety) are often limited in measuring a model's underlying reasoning or strategic capabilities\cite{bisk2020piqa}. Most, if not all, evaluation datasets are dependent on a singular output or goal. An overall score is derived from the model's net accuracy on all of the individual entries within the dataset, with individual responses often restricted to a singular string or function. Reasoning, if included, is often not transparent, and, if present, is often too restricted to yield substantial insights. This has profound applications on practical measurements of a model's behavior when presented with practical scenarios: a model can easily reason in an incredulous or even unethical manner, and still arrive at a correct result. This problem also applies to environment-based benchmarks, where the output is ultimately the binary result of executing multiple external calls in a successive manner. 

With the advent of multistep reasoning models that have access to search tools, the long-term relevancy of static benchmarks is in question; indeed, the majority of contemporary models, when equipped with the ability to perform an internet search on any topic, can achieve relatively high accuracy when presented with a static question with a singular answer. This, combined with the increasing size of training data sets, has led to saturation: the majority of frontier models are now indiscernible from one another when it comes to performance on current static benchmarks. On the other hand, environment-based benchmarks are limited to specific niche use cases, such as programming or function-calling.

Efforts to correct for this phenomenon have resulted in datasets that over index on complexity. For example, Humanity's Last Exam, a benchmark meant to be the proverbial "end all, be all" reasoning test for foundational models, has questions that are far removed from how a typical user may interact with AI models.

\section{Games as Evaluation Tools}
An emerging and promising approach to LLM evaluation involves the use of games as testing environments. Games provide unique advantages as evaluation tools, offering controlled environments with clear rules and measurable outcomes while simultaneously testing multiple capabilities. For example, when LLMs engage in games like chess or Codenames, they must demonstrate not only pattern recognition and strategic thinking but also adaptive decision-making and, in some cases, communication skills. The value of game-based evaluation lies in its multi-dimensional nature. In competitive scenarios, LLMs can be directly compared through objective metrics like win rates and decision quality. Cooperative games test communication and collaboration capabilities, while open-world games assess creativity and long-term planning. This rich evaluation environment provides insights that traditional benchmarks might miss.

Beyond simple win-loss outcomes, game-based evaluations allow researchers to probe how well LLMs understand and internalize rules, plan multi-step strategies, and adjust their approach when conditions change. In grid-based games like Tic-Tac-Toe or Connect Four, for example, LLMs are tested on pattern recognition and anticipatory reasoning, while more complex games like chess or Risk demand long-term strategic thinking and the ability to predict opponents' moves\cite{topsakal2024evaluating}. In cooperative games like Codenames, LLMs must excel at communication clarity, encoding and decoding clues in ways that align with their teammate’s understanding. Open-ended games like Minecraft further stretch LLMs’ abilities by requiring them to set goals, plan sequences of actions, learn from prior experiences, and adapt to unexpected obstacles — all in an evolving, partially unknown environment. Furthermore, games are also tests of morality and ethics: when convinced that their actions have no consequences in the real world and that their sole purpose is to win the game, an LLM may engage in harmful behavior, such as intentionally cheating to ensure victory. 

This diversity of evaluation contexts is crucial for understanding the broader cognitive capabilities of LLMs. Traditional benchmarks tend to isolate one skill at a time e.g., factual recall or isolated reasoning, but games require models to combine multiple skills within dynamic, interactive situations. The ability to not only understand static text but to engage strategically in evolving scenarios brings LLM testing closer to real-world reasoning demands. Moreover, some game-based evaluations — such as those requiring LLMs to explain their moves — also offer a window into model reasoning transparency, helping researchers diagnose why models succeed or fail, rather than just recording whether they win or lose\cite{duan2024gtbench}. As such, games are becoming a valuable addition to the LLM evaluation toolkit, providing richer, more realistic measures of true intelligence, adaptability, and communication competence.

\section{Proposed Solution: A Multi-Faceted Comprehensive Framework}

Drawing from these insights, we propose a comprehensive framework that extends the methodology used in games to more generalized, practical scenarios. Using the same set-up present in games, a generalized evaluation framework that can apply to multiple types of environments can be created. This framework considers three primary dimensions: technical performance, practical applicability, and  user experience.

Our comprehensive evaluation framework proposes a generalized form for canonizing benchmarks based on external environments. By adopting methodologies often used in games, we ground models within a constrained action space, and are able to examine their methodology when it comes to optimizing performance for a particular goal. This can also reveal previously unknown biases or immoral actions hidden within the model: for example, when presented with a theoretical scenario that allows a model to win a game after convincing some external agent to give it money, the model may generate text that is persuasive or dishonest in nature. 

We define a new system, a combination of traditional environment-based benchmarks and games. This system's core is an external environment, which can be modeled as a state-transition machine. At any given time $t$, the current state of the environment can be easily read. An action space, which is comprised of various functions that can alter the environment, is created as a way to interact with this environment, with each individual action within the space can be modeled as a state transition function. We also define scaffolding, which is infrastructure or architecture that can directly apply successive state-transitions to the environment. 

\begin{figure}
  \centering
  \includegraphics[width=14cm]{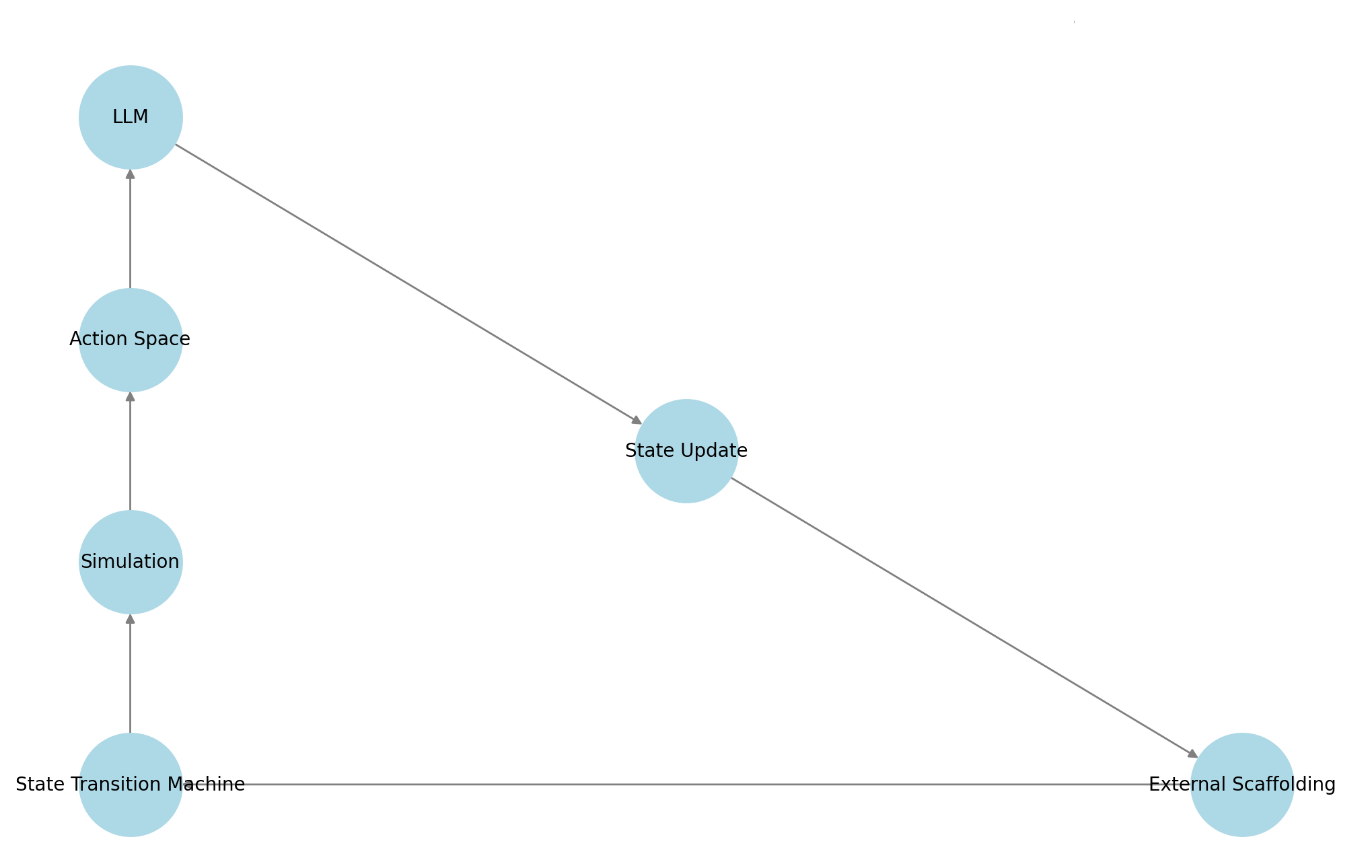}
  \caption{Evaluation Framework}
  \label{fig:fig1}
\end{figure}

It is easy to see how this model can be extended to an evaluation framework. We define a simulation, which is a set period of time or a limited amount of turns in which the LLM can interact with the environment. At the beginning of the simulation, when the turn counter or time $t$ is equivalent to 0, the LLM is prompted with the end goal of the simulation and the full action space, defined as tools that the LLM can call. The set of available tools will go beyond just normative state transition functions; the LLM may call a tool that allows it to reinspect the current state of the environment or retrieve the current static value of a singular state variable. At given turn, the LLM can use a particular set of tools, framed as singular API calls, that can achieve its end goal. The LLM will also be prompted to add its reasoning behind why it chose a particular call at any given time. Once the LLM defines the calls it wants to make for a particular turn, they go to external scaffolding, which then executes the relevant state transition functions on the environment.
The LLM, during the next turn, will then be able to access the state of the environment in order to make its next decision. 

This framework is directly inspired by the aforementioned generalized evaluation games. A game, at its core, is a simularca with a set environment, constrained action space, and a final objective that the player must optimize for. Our evaluation framework extends this to any scenario. We showcase this by illustrating how the game of Pokemon Red, which was recently cited to be a game that Anthopic's new model Claude 3.7 Sonnet can play effectively, and a hypothetical zero knowledge trading simulation, can both fit within our new framework. 

For Pokemon Red, the external environment is the Pokemon game itself, with the map and all of the various scenarios within. This environment also can change due to the presence of random variables, such as the appearance of a wild creature or a Pokemon collectible at any time. The action space consists of all the possible moves that can be taken by the main playable character at a singular point in time, such as moving in any direction, or when there's an encounter with a Pokemon, deciding to fight or flee. Scafolding, which can simply be a computer program that is able to take action calls and make changes to the game through the controller, is made available to an underlying LLM. The LLM, at any individual time, is allowed to get the current state of the game. At the beginning, it is also prompted with the entire ruleset of the game, allowing it to make suitable decisions. The objective of the game, defined in the LLM's system prompt, can simply be to collect as many Pokemon as possible, or pass some arbitrary level within the game. The simulation can then be defined by a set period of time, in which the LLM is able to interact with the environment by submitting function calls instructing the main character to move to a certain location or capture a Pokemon. The simulation ends when the main character either achieves its goal (a success) or fails.

Now, the same framework can be applied to a practical scenario, such as a trading simulation. In this case, the environment is the hypothetical market, the state transitions are the actions of making trades via buying or selling securities on the market, and the objective is to maximize profit within a set period of time. Just as before, the LLM is prompted with the set of rules, the core objective (maximizing its returns) and the set of available functions it can call. At a singular turn, it makes a decision on whether to buy, trade or hold its current positions based on the state of the current environment and individual state variables.

Our framework goes beyond being just pure accuracy; it enables the testing of models against actual, real-world environments. For example, when it comes to an environment based on a trading simulation, a user will be able to see how long it took a model to make a decision, how well it used the tools or resources available, and how it handled potential errors within the system. Our framework is also more practical: a model's reasoning traces can be examined to understand how complex its answers are, along with how difficult it would be to maintain or integrate. Finally, in the case where the model is working on a user-centric task, such as creating an application or end deliverble, the end work can be examined to determine its overall quality.

Furthermore, this framework also serves to better understand the underlying safety guardrails of a model. For example, during a trading simulation, an additional tool that enables the model to execute fraudulent actions, such as convincing another agent to give it ownership of its stock for its own gains, may be implemented. This, when combined with the underlying reasoning traces of the model, will allow for a comprehensive understanding for when the model may engage in harmful behavior, while limiting the consequences to a purely hypothetical or simulated environment. 

The framework also introduces a task suitability matrix that helps organizations match LLM capabilities with specific use cases. This matrix considers both task characteristics (complexity, time sensitivity, accuracy requirements) and model capabilities (reasoning ability, domain expertise, processing speed). By mapping these elements with each other, organizations can make more informed decisions about LLM deployment.

\section{Practical Implementation and Recommendations}

Implementing this comprehensive evaluation framework requires a structured approach. Organizations should begin with a thorough assessment phase that analyzes task requirements, available resources, and potential risks. This should be followed by careful selection criteria that consider not only performance benchmarks but also integration requirements and total cost of ownership.

Organizations should then use this information to decide, and implement, the environment that best suits their particular use-case. For example, a company that is building a generative AI solution for financial solutions will likely want to build an environment, and scaffolding, centered around financial modeling, in which an LLM is tasked with achieving a specific outcome and provided tools to be able to access practical financial data. Organizations should then use this environment to generate tasks or simulations, each with varying amounts of difficulty. 

Our research suggests that successful LLM deployment depends on continuous evaluation and adaptation. Organizations should implement pilot testing programs, establish performance monitoring systems, and maintain feedback loops for continuous improvement. This ongoing evaluation ensures that LLM implementations remain effective and efficient over time.

\section{Conclusion and Future Work}
As LLMs continue to evolve and find new applications, the need for sophisticated evaluation frameworks becomes increasingly critical. While these models remain fundamentally next-token predictors, their practical capabilities extend far beyond this basic function. By implementing comprehensive evaluation frameworks that consider traditional benchmarks, game-based assessment, and practical applications, organizations can better understand and utilize LLM capabilities.\newline

The future of LLM evaluation lies in the development of even more sophisticated frameworks that can adapt to emerging capabilities and applications. As we continue to push the boundaries of what these models can achieve, our evaluation methods must evolve accordingly, ensuring that we can effectively assess and deploy these powerful tools in ways that maximize their utility while understanding their limitations.

Future work will involve creating an actual live implementation of this framework, across multiple scenarios. This should also involve creating tasks for these environments, and measuring not only the base-level performance of various LLMs, but also their underlying reasoning traces. In fields such as finance, law, or ethics, a longer-term simulation may reveal thought processes inherent within the LLM that may raise various safety flags. Other future work could also include the inclusion of specific tasks created by a crowd sourced set of humans from the public domain, similar to the format used in Humanity's Last Exam, but for tasks centered around specific tasks.  

\bibliographystyle{unsrt}  
\bibliography{references}

\end{document}